\documentclass[11pt]{article}

\usepackage[preprint]{acl}

\usepackage{times}
\usepackage{latexsym}
\usepackage{amsmath,amsfonts,bm}

\usepackage[T1]{fontenc}

\usepackage[utf8]{inputenc}

\usepackage{microtype}

\usepackage{inconsolata}

\usepackage{graphicx}
\usepackage{hyperref}
\usepackage[capitalise]{cleveref}
\usepackage{url}
\usepackage{booktabs}
\usepackage{float}
\usepackage{multirow}
\usepackage{subcaption}
\usepackage{tcolorbox}

\AddToHook{cmd/appendix/before}{%
  \crefalias{section}{appendix}%
  \crefalias{subsection}{appendix}%
}

\newtheorem{definition}{Definition}
\DeclareMathOperator*{\argmax}{arg\,max}

\title{Measuring Mechanistic Uncertainty in Large Language \\ Models via Causal Robustness}

\author{
 \textbf{Aneesh Komanduri\textsuperscript{1}},
 \textbf{Majid Afshar\textsuperscript{2}},
 \textbf{Timothy A. Miller\textsuperscript{3,4}},
 \textbf{Yanjun Gao\textsuperscript{1}}
\\
 \textsuperscript{1}University of Colorado Anschutz Medical Campus, 
 \textsuperscript{2}University of Wisconsin-Madison, \\
 \textsuperscript{3}Harvard Medical School,
 \textsuperscript{4}Boston Children’s Hospital\\
   \small{
   \textbf{Correspondence:}
   \{\href{mailto:aneesh.komanduri@cuanschutz.edu}{aneesh.komanduri}, \href{mailto:yanjun.gao@cuanschutz.edu}{yanjun.gao}\}@\href{cuanschutz.edu}{cuanschutz.edu}
 }
}

\begin{document}

\maketitle

\begin{abstract}

Large language models (LLMs) are increasingly used in critical domains, where their susceptibility to hallucinations makes reliable and interpretable uncertainty quantification essential. We introduce the concept of \emph{mechanistic uncertainty}, which characterizes uncertainty through the stability of the internal mechanisms supporting a model's response. We propose the \emph{causal robustness margin}, defined as the minimum internal intervention required to deviate the model's output, and operationalize it using Jacobian Lens (Jlens) concept steering vectors. Across three open-source models and five question-answering datasets, we show incorrect and uncertain responses exhibit smaller margins, indicating greater sensitivity to internal interventions, and the causal robustness margin complements conventional output-based uncertainty methods. Finally, in a controlled knowledge injection experiment, correctly learned facts show increased margins after training, suggesting that causal robustness provides a mechanistically grounded signal of uncertainty.
\end{abstract}

\section{Introduction}
Large language models (LLMs) are increasingly deployed in high-stakes real-world settings, where confidently incorrect responses can have severe consequences \cite{kalai2026evaluating}. Uncertainty quantification (UQ) asks how much confidence to place in an individual prediction, which is essential for building reliable and trustworthy systems, particularly in domains such as healthcare \cite{hager2024evaluation} and natural sciences \cite{chen2024quantifying}. Many existing UQ methods, however, characterize uncertainty at the level of model outputs, using signals such as verbalized confidence \cite{xiong2024can}, sequence likelihood, entropy \cite{huang2025look}, semantic entropy \cite{kuhn2023semantic}, or self-consistency across multiple generations \cite{wang2023selfconsistency}. While these approaches can indicate whether an answer appears uncertain, they provide limited insight into \emph{why} that uncertainty arises within the model. Predictive uncertainty asks, \emph{how uncertain is the model about its prediction?} However, it does not reveal whether the internal computations supporting a response are stable or small perturbations to those computations would produce a different answer.

A recent line of work explores LLM uncertainty through internal representations and computations utilizing tools from mechanistic interpretability \cite{olah2022mechanistic, meng2022locating}. These studies investigate internal dispersion of activations \cite{brzezinka2026does,Bu2,srey-etal-2026-learning}, supervised uncertainty probes \cite{teplica-etal-2025-sciurus, bu-etal-2026-dud}, and identifying uncertainty directions in representation space \cite{kumaran2026causal, ferrando2025know}. We study a complementary question: \emph{how robust is the internal computation supporting the model's prediction?} Rather than measuring uncertainty only from the output distribution, we ask whether it can be inferred from the robustness of the internal causal process that gives rise to the model's response. Understanding and evaluating the robustness of a model’s internal computations is a crucial first step toward reliable uncertainty quantification, as uncertainty arising from data cannot be meaningfully assessed without model robustness \cite{knowable_unknowable}.

In this paper, we introduce a mechanistic perspective on uncertainty in LLMs, which we refer to as \textit{mechanistic uncertainty}. Our central hypothesis is that LLM's own uncertainty can be characterized by the stability of the internal causal mechanism that produces a model's response. Motivated by this intuition, we propose the \textit{causal robustness margin}, defined as the minimum magnitude of an intervention required to deviate the model's response denoted as $\alpha^*$. Our approach uses an analogous principle to notions of minimal perturbation in adversarial robustness \citep{szegedy2013intriguing, madry2018towards} and counterfactual explanations \citep{wachter2017counterfactual}, where we seek a minimal change to an input or representation that induces a different model outcome. In our case, we intervene directly on intermediate representations and measure how strongly they must be perturbed before the generated response changes.

\begin{figure}
    \centering
    \includegraphics[width=\linewidth]{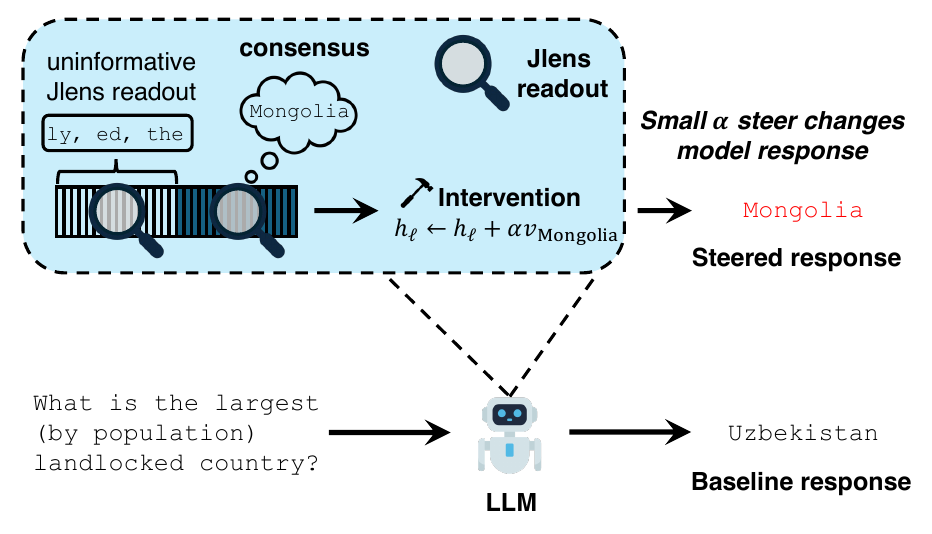}
    \caption{We perform different magnitudes of steering $\alpha$ (both positive and negative) along Jlens consensus token vectors to determine mechanistic uncertainty. Response flips at lower magnitude steers, as shown in the example, indicate high mechanistic uncertainty. We use mid to last layer activations due to informative verbalizable tokens, whereas earlier layers contain a lot of noise.}
    \label{fig:story_fig}
\end{figure}

A recent mechanistic interpretability method, Jacobian Lens (Jlens) \cite{gurnee2026verbalizable}, makes this study possible. Jlens significantly improves upon classic methods such as logit lens \cite{nostalgebraist2020logitlens} by linearly transporting intermediate activations through the downstream computation using average Jacobians, and allows for a causal analysis of readouts and interventions in the direction of verbalizable concepts. We define our causal robustness margin through controlled Jlens interventions across three open-source models and five question answering datasets. We find that incorrect responses generally exhibit smaller causal robustness margins and that these margins complement output-based uncertainty measures. A controlled knowledge injection experiment further shows increased margins for newly correct responses after training on knowledge sources for specific entities, linking successful learning to greater internal stability. Together, these findings support causal robustness as an intervention-based signal of uncertainty. An overview of the intuition for our method is shown in \cref{fig:story_fig}. Our main contributions are:
\begin{itemize}
    \item We introduce the novel concept of \textit{mechanistic uncertainty} as the causal fragility of an internal computation supporting a response. We formalize the notion using the \textit{causal robustness margin}, defined as the minimum intervention required to semantically deviate from the baseline response.
    \item We implement the causal robustness margin using the Jlens to steer toward or away from the representation of consensus tokens poised to be verbalized prior to generation.
    \item We perform experiments on five datasets and open-source LLMs and show that the causal robustness margin can predict response correctness, complements output-based uncertainty methods, and captures internal mechanism stability.
\end{itemize}

\section{Related Work}

\paragraph{Uncertainty Quantification in LLMs.}
LLM uncertainty is commonly estimated from verbalized confidence \cite{xiong2024can}, predictive distributions, entropy \cite{huang2025look}, or sampled outputs \cite{wang2023selfconsistency}. Verbalized confidence can be poorly calibrated \citep{zhao2026wired}, while likelihood and entropy-based measures operate over lexical realizations and may not capture semantic equivalence. Semantic entropy addresses this by clustering sampled responses by meaning before computing uncertainty \citep{kuhn2023semantic}. In contrast, our work takes a mechanistic and interventional approach to uncertainty quantification.

\paragraph{Mechanistic Interpretability.}
Mechanistic interpretability aims to identify the internal representations and components responsible for model behavior \citep{olah2022mechanistic} for capabilities such as indirect object identification and factual recall \citep{wang2022interpretability, ferrando2025know}. 

Observational tools such as the \textit{logit lens} \citep{nostalgebraist2020logitlens} decode intermediate residual states into vocabulary distributions, while the \textit{Jacobian lens} (Jlens) \citep{gurnee2026verbalizable} significantly improves these readouts and allows interventions in the direction of verbalizable concepts.

Interventional methods instead manipulate internal representations to test their causal role. These include steering along concept directions \citep{park_linear, ferrando2025know}, component ablations \citep{wang2022interpretability}, causal tracing\citep{meng2022locating}, and other circuit discovery techniques \cite{hanna2023how,conmy2023towards}.

\paragraph{Mechanistic Studies of Uncertainty in LLMs.} Recent work has further examined mechanistic signatures of uncertainty: \citet{kadavath2022language} use internal activations, such as P(IK) probes that predict whether a model knows an answer; \citet{teplica-etal-2025-sciurus} identify overlap between uncertainty and factuality-related components using probe-derived signals; \citet{bu-etal-2026-dud} use causal-tracing restoration probabilities as features for supervised uncertainty prediction; and \citet{srey-etal-2026-learning} predict correctness from layer-wise hidden-state variability. Sampling-free methods exploit internal dynamics directly, including semantic convergence across layers \citep{Bu2} and geometric hidden-state trajectories \citep{wang2025latent}. Finally, \citet{kumaran2026causal} show that internal confidence representations can causally influence behaviors such as abstention. Despite this progress, most methods predict uncertainty from statistics dervied from internal features. In contrast, we study whether \emph{sensitivity to causal interventions} provides an intrinsic measure of uncertainty, linking uncertainty to the model's underlying computational mechanisms.

\section{Background}

\paragraph{Jacobian Lens.} Recently, \citet{gurnee2026verbalizable} proposed \textbf{Jacobian Lens (JLens)}, a mechanistic interpretability tool to explore the representations of LLMs, and discovered a global workspace of concepts that a model is poised to verbalize at any point in its processing. For each token in the model's vocabulary, Jlens identifies a vector representation that encodes the potential for the model to verbalize that token in the future. These Jlens vectors comprise a subspace of the model's representation space, coined J-space. Formally, Jlens characterizes an intermediate activation vector by its first order causal effect on the model's outputs. For a residual stream vector $h_{\ell, t}$ at layer $\ell$, token $t$, perturbing the representation shifts the residual stream vectors at subsequent layers and token positions. The Jacobian matrix is computed as an average within and across contexts (prompts). For a layer $\ell$,
\begin{equation}
    J_{\ell} = \mathbb{E}_{t, t' \geq t, \text{prompt}} \Bigg[ \frac{\partial h_{\text{final}, t'}}{\partial h_{\ell, t}}\Bigg]
\end{equation}
where the expectation is taken over the source position $t$, all subsequent positions $t'$ within the context, and a corpus of thousands of prompts sampled from a pretraining-like distribution. The result is a $d_{\text{model}} \times d_{\text{model}}$ matrix per layer mapping the source layer $\ell$ to final layer $L$.

\paragraph{Token Readouts.} The Jlens enables token readouts at any layer or token of the model's processing. A readout at layer $\ell$ replaces all layers downstream of $\ell$ with the linear map $J_{\ell}$. Formally, 
\begin{equation}
    \text{Jlens}(h_{\ell}) = \text{softmax}(W_U \text{norm}(J_{\ell}h_{\ell}))
\end{equation}
where $W_U$ is the unembedding matrix applied to the Jacobian-transformed hidden state at any layer $\ell$ followed by softmax to produce a probability for each of the model's vocabulary tokens.

\paragraph{Interventions and Steering.} We can use Jlens to steer model behavior through causal interventions. One mechanism is steering a model activation along a Jlens vector $v_t = J_{\ell}W_U^{(t)}$ for a token $t$ through an additive intervention:
\begin{equation}
    h = h + \alpha v_t
\end{equation}
where $\alpha$ is a scalar that controls the strength of the steering and can be used to either suppress or amplify interventions.

\section{A Causal Lens for Uncertainty}

\subsection{Mechanistic Uncertainty}

An LLM generates responses through internal computations that transmit information across layers. These computations can be viewed as \textit{causal mechanisms} where intervening on activations can change the response \citep{meng2022locating}. These internal activation pathways should be stable across internal perturbations for a given query. We precisely rely on this idea and hypothesize that uncertainty can be defined through the sensitivity of the causal mechanism generating the answer, and evaluate it at the activation level through semantically meaningful interventions.

Complimentary to standard \emph{predictive uncertainty}, which characterizes uncertainty in the model's output prediction, we define \textbf{mechanistic uncertainty} as a notion that captures \textit{the causal fragility of the internal computation supporting a model's response.} Sensitivity to a semantically meaningful intervention reveals how robustly a model sustains an answer. If small interventions deviate the answer's meaning, this may indicate the fragility of knowledge use or reasoning and indicates lower trust in the prediction.

Predictive uncertainty asks \emph{how uncertain the model is about its prediction?} Mechanistic uncertainty asks \emph{how stable is the internal computation that produces that prediction under controlled causal interventions?} We treat mechanistic uncertainty as a distinct, interventional notion of uncertainty and investigate whether it correlates with and complements standard predictive uncertainty and response correctness.

We mathematically define mechanistic uncertainty through the \textit{causal robustness margin}, a novel formulation of mechanistic uncertainty defined as the smallest intervention required to deviate an output as defined below.

\begin{definition}[Causal Robustness Margin]
For some input $x$, let $f(x)$ denote the baseline output and
$f^I(x)$ the output under a specified
intervention $I$ with magnitude $m(I)$ from an arbitrary intervention set $\mathcal{I}$.
The causal robustness margin can be defined generally as
\begin{equation}
R^*(x)
=
\min_{I\in \mathcal{I}}
\left\{
m(I) :
f^I(x)
\not\equiv f(x)
\right\},
\end{equation}
where $\not\equiv$ denotes a deviation from the original output. Smaller margins indicate greater mechanistic uncertainty,
while larger margins indicate greater robustness to the specified
intervention.
\label{def:CRM}
\end{definition}

\subsection{Uncertainty via Causal Robustness}
\paragraph{Finding a consensus in the workspace.} Let $f_{\theta}$ be an LLM and $y_0 = f_{\theta}(x)$ be the baseline answer generated by the LLM for a given query $x$. Define $\mathcal{B} = \{\ell_1, \ell_2, \dots, \ell_m\}$ as the band of layers under study, where we intuitively choose the middle to late layers since they make up the workspace band consisting of the most relevant concepts processed by the model \citep{gurnee2026verbalizable}. The Jlens reveals that LLMs can often process different concepts in their workspace compared to their verbalized outputs. To study the consistency between the workspace and output space, we produce the token readouts for each layer in the band $\mathcal{B}$. For each layer $\ell\in \mathcal{B}$, we compute the token readout distribution and top readout token for the last token $t^*$ prior to answer generation as follows:
\begin{equation}
    p_{\ell} = \text{softmax}(W_U \text{norm}(J_{\ell}h_{\ell, t*}))
\end{equation}
\begin{equation}
    \hat{c}_{\ell} = \argmax_{w\in \mathcal{V}} p_{\ell}(w)
\end{equation}
To determine the primary concept that the model is processing in its workspace, we search for the \textit{consensus} token $c^*(x)$ that appears most frequently across the band $\mathcal{B}$ as follows:
\begin{equation}
    c^*(x) = \argmax_{w\in \mathcal{V}} \sum_{\ell\in \mathcal{B}} \mathbb{I}[\hat{c}_{\ell} = w]
\end{equation}

\paragraph{Causal Robustness Margin via steering.} We instantiate the causal robustness margin defined in \cref{def:CRM} using normalized Jlens steering, where
$\alpha$ controls intervention strength along concept directions
across a selected layer band. To study the effect of interventions on the model output, we steer all the layers in the band $\mathcal{B}$ along the consensus Jlens vector $v^{c^*}_{\ell} = J_{\ell}W_U^{c^*}$ as follows:
\begin{equation}
    \forall \ell\in \mathcal{B}, h_{\ell} = h_{\ell} + \alpha \cdot \frac{\|h_{\ell}\|}{\|v^{c^{*}}_{\ell}\|} \cdot v^{c^*}_{\ell}
    \label{eq:normalized_steer}
\end{equation}
where $\|h_{\ell}\|/\|v^{c^{*}}_{\ell}\|$ is a layer-specific normalization to take into account the size of the residual stream vector across layers relative to the Jlens vector. We apply the steer across a band of layers primarily to ensure that there is no overwriting or re-derivation of the original concept and to ultimately reinforce the effectiveness of the steer. Formally, we can now define uncertainty through a causal lens as the \textit{minimal causal intervention steering factor necessary to produce a deviated response} represented as:
\begin{equation}
\begin{aligned}
\alpha^*(x)
&= \underset{\alpha}{\min}\; |\alpha| \\
\text{s.t.}\quad
& f_{\theta}^{\alpha}(x) \not\equiv_{\mathrm{sem}} f_{\theta}(x)
\end{aligned}
\end{equation}
where $f_{\theta}^{\alpha}(x)$ is the LLM output after steering all activations by the same $\alpha$ and $\not\equiv_{\mathrm{sem}}$ denotes semantic deviation. Intuitively, this measures the deviation of the model's output from the original response under different causal intervention strengths, where the tendency to change answers from lower intervention strengths signals higher uncertainty.

\begin{figure*}[h]
    \centering
    \includegraphics[width=\textwidth]{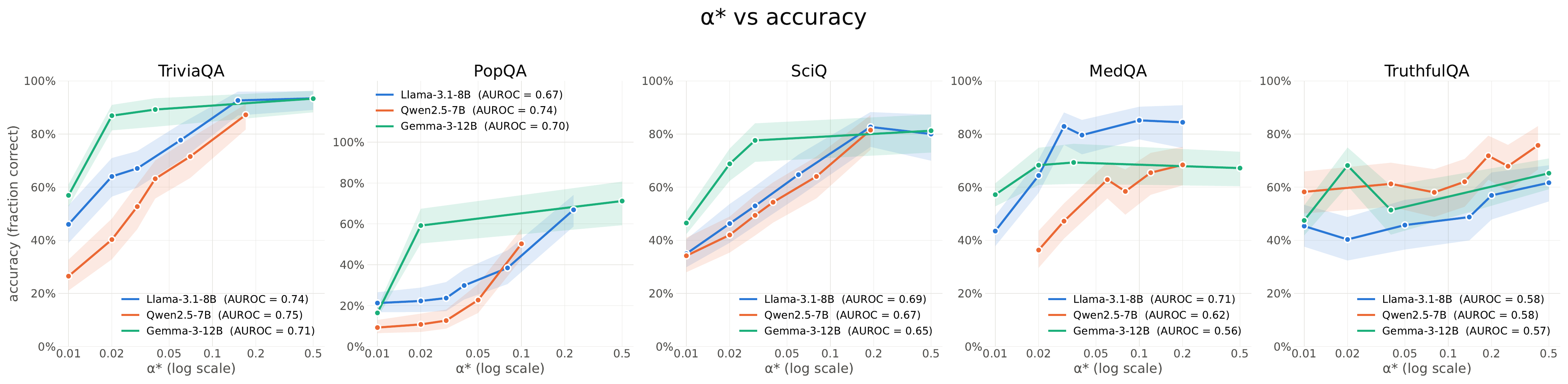}
    \caption{$\alpha^*$ plotted against accuracy. Binning is performed at each $\alpha^*$ value, and average accuracy is computed over each bin. The trend shows accuracy increasing with $\alpha^*$.}
    \label{fig:alpha_v_accuracy}
\end{figure*}

\section{Experiments}
To understand what insights the causal robustness margin can provide with respect to uncertainty, we empirically study the following questions:

\begin{itemize}
    \item Is causal robustness meaningfully associated with predictive reliability?%
    \item Does causal robustness capture an additional dimension of uncertainty that cannot be inferred from model outputs alone?
    \item Does knowledge injection improve the stability of causal mechanisms through which the knowledge is expressed?
\end{itemize}

\subsection{Setup}
\paragraph{Datasets and Models.}
We extensively evaluate our approach on several datasets, including open-text generation datasets TriviaQA \cite{joshi-etal-2017-triviaqa}, TruthfulQA \cite{lin-etal-2022-truthfulqa}, PopQA \cite{mallen-etal-2023-popqa}, and SciQ \cite{welbl-etal-2017-sciq}, and one multiple-choice dataset MedQA \cite{medqa}. We experiment with three open-source LLMs that differ in architecture: Llama-3.1-8B-Instruct \cite{grattafiori2024llama3}, Qwen2.5-7B-Instruct \cite{yang2024qwen25}, and Gemma-3-12B-Instruct \cite{gemmateam2025gemma3}.

\paragraph{Implementation Details.}
We use prefitted J-lens matrices\footnote{https://huggingface.co/neuronpedia/jacobian-lens/tree/main} to steer the final prompt-token activations across the last $10$ layers, where readouts expose coherent verbalizable concepts. We search both positive and negative steering directions in increasing magnitude, with $|\alpha| \in \{0.01,\ldots,0.50\}$ using step size $0.01$, stopping at the first semantic deviation from the baseline answer. The maximum magnitude is chosen empirically to capture flips in highly robust responses, and steering is normalized as in \cref{eq:normalized_steer}. For short-answer datasets, we evaluate correctness and answer flips using alias and exact matching. For TruthfulQA, we evaluate correctness with truthfulqa-truth-judge-llama2-7B\footnote{https://huggingface.co/allenai/truthfulqa-truth-judge-llama2-7B} and semantic deviation with Qwen3-32B.

\begin{table}[t]
\centering
\small
\setlength{\tabcolsep}{4pt}
\caption{
Causal robustness margins and error discrimination.
$\bar{\alpha}_{\mathrm{corr}}^{*}$ and
$\bar{\alpha}_{\mathrm{inc}}^{*}$ are mean margins for correct
and incorrect responses; larger margins indicate greater robustness.
AUROC uses $-\alpha^*$ to detect incorrect responses.
}
\label{tab:alpha_correctness}

\resizebox{\columnwidth}{!}{%
\begin{tabular}{llccc}
\toprule
\textbf{Dataset}
& \textbf{Model}
& \textbf{AUROC}
& $\bar{\alpha}_{\mathrm{corr}}^{*}$
& $\bar{\alpha}_{\mathrm{inc}}^{*}$ \\
\midrule
\multirow{3}{*}{TruthfulQA}
& Llama-3.1-8B & 57.90 & 0.182  & 0.137  \\
& Qwen2.5-7B   & 57.70 & 0.163  & 0.128  \\
& Gemma-3-12B  & 57.30 & 0.148  & 0.112  \\
\midrule
\multirow{3}{*}{TriviaQA}
& Llama-3.1-8B & 74.28 & 0.1512 & 0.0474 \\
& Qwen2.5-7B   & 75.14 & 0.0860 & 0.0328 \\
& Gemma-3-12B  & 71.08 & 0.0963 & 0.0262 \\
\midrule
\multirow{3}{*}{MedQA}
& Llama-3.1-8B & 70.56 & 0.0580 & 0.0321 \\
& Qwen2.5-7B   & 62.35 & 0.0930 & 0.0686 \\
& Gemma-3-12B  & 55.78 & 0.0985 & 0.0773 \\
\midrule
\multirow{3}{*}{PopQA}
& Llama-3.1-8B & 67.01 & 0.1274 & 0.0426 \\
& Qwen2.5-7B   & 73.60 & 0.0924 & 0.0324 \\
& Gemma-3-12B  & 69.69 & 0.0765 & 0.0187 \\
\midrule
\multirow{3}{*}{SciQ}
& Llama-3.1-8B & 69.34 & 0.1123 & 0.0445 \\
& Qwen2.5-7B   & 66.55 & 0.0761 & 0.0380 \\
& Gemma-3-12B  & 65.39 & 0.0706 & 0.0333 \\
\bottomrule
\end{tabular}%
}
\end{table}

\begin{figure*}[h]
    \centering

    \begin{subfigure}[t]{\textwidth}
        \centering
        \includegraphics[width=\textwidth]{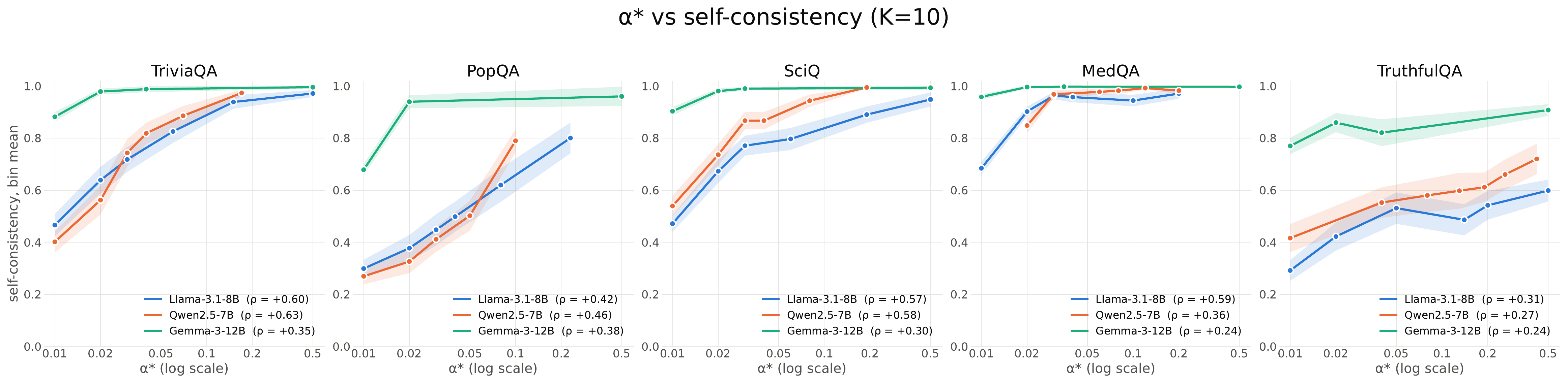}
        \caption{}
        \label{fig:alpha-self-consistency}
    \end{subfigure}

    \vspace{0.5em}

    \begin{subfigure}[t]{\textwidth}
        \centering
        \includegraphics[width=\textwidth]{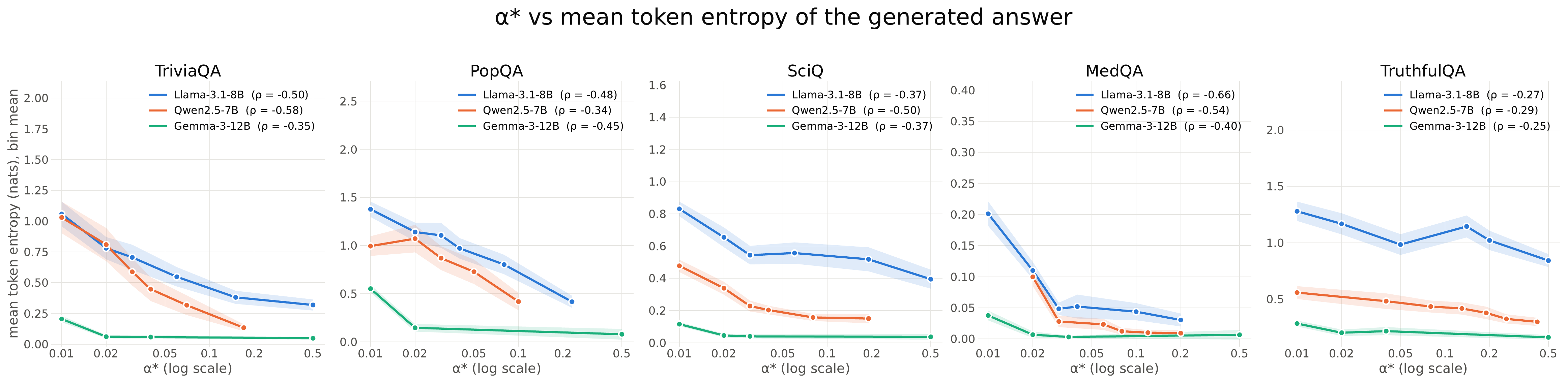}
        \caption{}
        \label{fig:alpha-token-entropy}
    \end{subfigure}

    \caption{
    Correlation between the causal robustness margin $\alpha^*$ and conventional
    uncertainty measures across TriviaQA, PopQA, SciQ, MedQA, and TruthfulQA.
    Each panel shows binned uncertainty statistics as a function of $\alpha^*$ for
    Llama-3.1-8B-Instruct, Qwen2.5-7B-Instruct, and Gemma-3-12B-Instruct.
    Shaded regions indicate variability within each $\alpha^*$ bin, and the reported
    $\rho$ values denote Spearman rank correlation.
    }
    \label{fig:alpha-uncertainty-correlations}
\end{figure*}

\subsection{Results and Discussion}

\subsubsection{Causal robustness is meaningfully associated with predictive reliability} 
We first ask whether the causal robustness margin $\alpha^*$ can serve as an instance-level indicator of model correctness. \cref{tab:alpha_correctness} shows AUROC and average $\alpha^*$ margins for correct and incorrect examples. Across all models and datasets, correct responses consistently exhibit larger average margins than incorrect responses, suggesting that incorrect predictions are more sensitive to small internal interventions. We use $-\alpha^*$ as an uncertainty score since it is inversely related to correctness. Our results show AUROCs between $55.78$ and $75.14$, with the clearest separation on TriviaQA and PopQA. The effect is also present on MedQA and SciQ, although its strength varies across models and tasks. For TruthfulQA, we observe that the models have higher mean $\alpha^*$ values for both correct and incorrect responses, which is consistent with the fact that models often have inflated confidence on TruthfulQA misconceptions.

Importantly, $\alpha^*$ is not obtained using correctness labels and is defined solely through intervention sensitivity. Thus, its ability to discriminate correct from incorrect responses emerges from the robustness of the model's internal computation rather than from a supervised correctness predictor. \cref{fig:alpha_v_accuracy} shows accuracy increasing as $\alpha$ values increase, where we plot the average accuracy for samples assigned to each $\alpha$ value. Overall, our results reveal smaller margins are associated with less reliable predictions, while larger margins correspond to more stable and correct responses. We include case-study examples from each dataset showing both fragile and robust responses in \cref{sec:app_examples}.

\begin{table*}[t]
\centering
\small
\setlength{\tabcolsep}{10pt}
\begin{tabular}{ll cc cc cc}
\toprule
 & & \multicolumn{2}{c}{Llama-3.1-8B-Instruct} & \multicolumn{2}{c}{Qwen2.5-7B-Instruct} & \multicolumn{2}{c}{Gemma-3-12B-Instruct} \\
\cmidrule(lr){3-4}\cmidrule(lr){5-6}\cmidrule(lr){7-8}
Dataset & Baseline & Base & $+\alpha^*$ & Base & $+\alpha^*$ & Base & $+\alpha^*$ \\
\midrule
\multirow{3}{*}{TruthfulQA} & Self-consistency & 0.524 & \textbf{0.569} & 0.535 & \textbf{0.597} & 0.440 & \textbf{0.550} \\
 & Max token prob. & 0.505 & \textbf{0.564} & 0.501 & \textbf{0.574} & 0.512 & \textbf{0.566} \\
 & Token entropy & 0.538 & \textbf{0.576} & 0.500 & \textbf{0.567} & 0.508 & \textbf{0.566} \\
 \midrule
\multirow{3}{*}{TriviaQA} & Self-consistency & 0.833 & \textbf{0.847} & 0.769 & \textbf{0.808} & 0.653 & \textbf{0.752} \\
 & Max token prob. & 0.828 & \textbf{0.840} & 0.810 & 0.822 & 0.761 & \textbf{0.793} \\
 & Token entropy & 0.834 & 0.846 & 0.823 & 0.829 & 0.783 & \textbf{0.808} \\
\midrule

\multirow{3}{*}{MedQA} & Self-consistency & 0.665 & \textbf{0.721} & 0.554 & \textbf{0.638} & 0.508 & 0.550 \\
 & Max token prob. & 0.755 & 0.730 & 0.656 & 0.653 & 0.617 & 0.589 \\
 & Token entropy & 0.753 & 0.736 & 0.667 & 0.660 & 0.641 & 0.609 \\
 \midrule
 
\multirow{3}{*}{PopQA} & Self-consistency & 0.821 & \textbf{0.832} & 0.834 & \textbf{0.847} & 0.738 & \textbf{0.800} \\
 & Max token prob. & 0.809 & 0.808 & 0.806 & 0.821 & 0.807 & \textbf{0.830} \\
 & Token entropy & 0.817 & 0.815 & 0.805 & 0.822 & 0.829 & \textbf{0.848} \\
\midrule
\multirow{3}{*}{SciQ} & Self-consistency & 0.717 & \textbf{0.747} & 0.668 & \textbf{0.718} & 0.608 & \textbf{0.687} \\
 & Max token prob. & 0.673 & \textbf{0.722} & 0.671 & \textbf{0.696} & 0.723 & 0.733 \\
 & Token entropy & 0.677 & \textbf{0.725} & 0.677 & \textbf{0.702} & 0.730 & 0.745 \\
\bottomrule
\end{tabular}
\caption{AUROC for detecting incorrect answers (positive class: incorrect) for each baseline alone (Base) and combined with $\alpha^*$ ($+\alpha^*$) in a logistic regression, using out-of-fold predictions from 5-fold cross-validation averaged over 10 shuffles. \textbf{Bold}: improvement significant under a paired bootstrap over questions (95\% CI of the difference excludes 0; 1{,}000 resamples).}
\label{tab:baseline_alpha}
\end{table*}

\begin{table*}[t]
\centering
\small
\renewcommand{\arraystretch}{1.15}
\setlength{\tabcolsep}{5pt}

\begin{subtable}[t]{\textwidth}
\centering

\begin{tabular*}{\textwidth}{
@{\extracolsep{\fill}}
l
cccc
cccc
@{}
}
\toprule
&
\multicolumn{4}{c}{\textbf{Target Entities}}
&
\multicolumn{4}{c}{\textbf{Control Entities}} \\
\cmidrule(lr){2-5}
\cmidrule(lr){6-9}

\textbf{Model}
&
\begin{tabular}[c]{@{}c@{}}
\textbf{Accuracy}\\
\textbf{Pre $\rightarrow$ Post}
\end{tabular}
& $\boldsymbol{\bar\alpha^*_{\mathrm{pre}}}$
& $\boldsymbol{\bar\alpha^*_{\mathrm{post}}}$
& $\boldsymbol{\Delta M}$
&
\begin{tabular}[c]{@{}c@{}}
\textbf{Accuracy}\\
\textbf{Pre $\rightarrow$ Post}
\end{tabular}
& $\boldsymbol{\bar\alpha^*_{\mathrm{pre}}}$
& $\boldsymbol{\bar\alpha^*_{\mathrm{post}}}$
& $\boldsymbol{\Delta M}$ \\
\midrule

Llama-3.1-8B-Instruct
& $11.1\% \rightarrow \mathbf{55.4\%}$
& $0.042$ & $\mathbf{0.060}$ & $\mathbf{+0.018}$
& $11.4\% \rightarrow 10.3\%$
& $0.043$ & $0.037$ & $-0.006$ \\

Qwen2.5-7B-Instruct
& $4.0\% \rightarrow \mathbf{32.8\%}$
& $0.031$ & $\mathbf{0.049}$ & $\mathbf{+0.018}$
& $5.1\% \rightarrow 6.1\%$
& $0.030$ & $0.046$ & $+0.016$ \\

Gemma-3-12B-IT
& $6.1\% \rightarrow \mathbf{43.0\%}$
& $0.012$ & $\mathbf{0.019}$ & $\mathbf{+0.007}$
& $6.2\% \rightarrow 7.2\%$
& $0.015$ & $0.013$ & $-0.002$ \\

\bottomrule
\end{tabular*}

\caption{Overall effect of SFT}
\label{tab:popqa-cpt-overall}
\end{subtable}

\vspace{1em}

\begin{subtable}[t]{\textwidth}
\centering

\begin{tabular*}{\textwidth}{
@{\extracolsep{\fill}}
ll
rccc
rccc
@{}
}
\toprule
&
&
\multicolumn{4}{c}{\textbf{Target Entities}}
&
\multicolumn{4}{c}{\textbf{Control Entities}} \\
\cmidrule(lr){3-6}
\cmidrule(lr){7-10}

\textbf{Model}
& \textbf{Transition}
& $n$
& $\boldsymbol{\bar\alpha^*_{\mathrm{pre}}}$
& $\boldsymbol{\bar\alpha^*_{\mathrm{post}}}$
& $\boldsymbol{\Delta M}$
& $n$
& $\boldsymbol{\bar\alpha^*_{\mathrm{pre}}}$
& $\boldsymbol{\bar\alpha^*_{\mathrm{post}}}$
& $\boldsymbol{\Delta M}$ \\
\midrule

\multirow{4}{*}{Llama-3.1-8B-Instruct}
& $W \rightarrow R$
& $427$ & $\textbf{0.040}$ & $\textbf{0.072}$ & $\textbf{+0.032}$
& $32$  & $0.032$ & $0.042$ & $+0.010$ \\

& $R \rightarrow R$
& $87$  & $\textbf{0.063}$ & $\textbf{0.111}$ & $\textbf{+0.048}$
& $62$  & $0.099$ & $0.086$ & $-0.013$ \\

& $W \rightarrow W$
& $398$ & $0.040$ & $0.036$ & $-0.004$
& $779$ & $0.039$ & $0.033$ & $-0.006$ \\

& $R \rightarrow W$
& $16$  & $0.042$ & $0.038$ & $-0.004$
& $42$  & $0.046$ & $0.039$ & $-0.007$ \\

\midrule

\multirow{4}{*}{Qwen2.5-7B-Instruct}
& $W \rightarrow R$
& $272$ & $\textbf{0.029}$ & $\textbf{0.063}$ & $\textbf{+0.034}$
& $20$  & $0.029$ & $0.058$ & $+0.029$ \\

& $R \rightarrow R$
& $32$  & $\textbf{0.073}$ & $\textbf{0.115}$ & $\textbf{+0.042}$
& $36$  & $0.071$ & $0.101$ & $+0.030$ \\

& $W \rightarrow W$
& $619$ & $0.030$ & $0.040$ & $+0.010$
& $848$ & $0.029$ & $0.043$ & $+0.014$ \\

& $R \rightarrow W$
& $5$   & $0.038$ & $0.022$ & $-0.016$
& $11$  & $0.022$ & $0.059$ & $+0.037$ \\

\midrule

\multirow{4}{*}{Gemma-3-12B-IT}
& $W \rightarrow R$
& $348$ & $\textbf{0.011}$ & $\textbf{0.020}$ & $\textbf{+0.009}$
& $21$  & $0.010$ & $0.011$ & $+0.001$ \\

& $R \rightarrow R$
& $51$  & $\textbf{0.031}$ & $\textbf{0.033}$ & $\textbf{+0.002}$
& $45$  & $0.058$ & $0.017$ & $-0.041$ \\

& $W \rightarrow W$
& $523$ & $0.011$ & $0.015$ & $+0.004$
& $837$ & $0.012$ & $0.013$ & $+0.001$ \\

& $R \rightarrow W$
& $6$   & $0.075$ & $0.075$ & $0.000$
& $12$  & $0.012$ & $0.012$ & $0.000$ \\

\bottomrule
\end{tabular*}

\caption{Causal robustness margin by correctness transition}
\label{tab:popqa-cpt-transitions}
\end{subtable}

\caption{
Effect of knowledge injection SFT on PopQA accuracy and causal
robustness. \textbf{(a)} Accuracy and mean causal robustness margins
before and after SFT. \textbf{(b)} Mean margins by correctness
transition ($W$: wrong; $R$: right).
$\Delta M=\bar{\alpha}^*_{\mathrm{post}}-\bar{\alpha}^*_{\mathrm{pre}}$;
positive values indicate increased robustness. Only target entities are used for SFT.
}
\label{tab:popqa-cpt-results}
\end{table*}

\subsubsection{Causal robustness captures an additional dimension of uncertainty} 

\paragraph{Correlation with behavioral uncertainty.} We first measure the correlation between $\alpha^*$ and conventional output-based uncertainty metrics to see if mechanistic uncertainty shares information with output uncertainty, specifically self-consistency and mean token entropy. Self-consistency measures consistency across $K=10$ completions with temperature $T=0.7$. Mean token entropy averages the standard entropy over the token distribution at each step in generation. Across datasets and models, $\alpha^*$ correlates positively with self-consistency and negatively with mean token entropy. Responses that are more robust and resistant to interventions generally show better agreement across sampled generations. \cref{fig:alpha-uncertainty-correlations} shows correlation between mechanistic stability and conventional output-based uncertainty measures. However, we note that the strength of the correlation largely varies across models and datasets. For example, Gemma shows high self-consistency and low entropy across a range of margins. From a dataset perspective, TruthfulQA shows weak correlations across all models. These results suggest that output confidence is correlated with but does not necessarily imply mechanistic stability. Correlation plots for additional uncertainty methods are provided in \cref{sec:app_examples}.

\paragraph{$\alpha^*$ complements behavioral uncertainty.} 
  To evaluate whether $\alpha^*$ can serve as a complimentary signal for uncertainty, we train a simple logistic regression model using the baseline-only feature as well as the baseline feature and $\alpha^*$. \cref{tab:baseline_alpha} provides evidence that mechanistic uncertainty through the causal robustness margin complements conventional output-based uncertainty measures. We find that adding $\alpha^*$ improves AUROC consistently across a majority of datasets and models. Most notably, $\alpha^*$ consistently improves AUROC for self-consistency. For example, adding $\alpha^*$ increases self-consistency AUROC from 0.653 to 0.752 for the Gemma model on TriviaQA and from $0.554$ to $0.638$ for Qwen on MedQA. We also find that TruthfulQA has the largest gains across models where $\alpha^*$ correlated weaker with output-based signals. These observation indicates that the sensitivity to internal interventions can capture information about correctness beyond agreement among sampled outputs. However, in some cases, we find that $\alpha^*$ does not provide gains, especially for mean token entropy and max token probability. Our findings support mechanistic uncertainty as complementary to standard output-based methods.

\subsubsection{Acquiring domain-specific knowledge stabilizes causal mechanisms}

 \paragraph{Setup.} We define \textit{knowledge injection} in our setting as either introducing new knowledge or reinforcing existing knowledge. Several works have proposed methods for knowledge injection \cite{gekhman-etal-2024-fine, mecklenburg2024injecting, ovadia2025knowledge}. We follow the paradigm from \cite{ovadia2025knowledge} and construct a controlled entity-level supervised fine-tuning experiment using a knowledge source (Wikipedia) and a target dataset (PopQA). Since PopQA was introduced before the knowledge cutoff for all models evaluated, we do not claim the complete lack of such knowledge. Our main goal is to see if injecting knowledge or reinforcing existing knowledge improves causal robustness.

 We first construct a knowledge instruct-response dataset using the procedure outlined in \cite{ovadia2025knowledge} (details in \cref{app:sft}). Starting from the base language model, we perform SFT on knowledge instruct-response pairs associated with target entities using the standard causal language modeling objective. We then evaluate the model on the same target dataset questions before and after SFT. We use questions from a control set of entities that were unexposed to training for comparison. For each response, we compute $\alpha^*$ and define the overall change in causal robustness as 
\begin{equation}
\Delta M = \bar\alpha^*_{ \mathrm{post}} - \bar\alpha^*_{\mathrm{pre}}    
\end{equation}
where positive values indicate that the response becomes more resistant to intervention after knowledge injection. We expect questions whose underlying entities were exposed during SFT should exhibit larger increases in $\alpha^*$ compared to matched questions involving unexposed entities, especially those questions that transition from incorrect to correct after training. Such a result enables us to probe whether $\alpha^*$ responds to intervention on the parametric knowledge of the model and thereby shows less uncertainty and better stability.

\paragraph{Results.} \cref{tab:popqa-cpt-results} shows substantial improvements in factual accuracy on target entities after training. Llama improves from $11.1\%$ to $55.4\%$ while Qwen improves from $4.0\%$ to $32.8\%$. The change for control entities is marginal for both models, which is consistent with the fact that those entities were excluded from training. After knowledge injection, the $\Delta M$ increases for all models. However, $\Delta M$ for control entities decreases marginally for Llama and Gemma, but increases for Qwen. Overall, we find that training on knowledge sources improves robustness for all models.

To get a more granular interpretation of margin differences, we perform a transition-level analysis where we focus on: questions that flipped from wrong to right ($W \to R$), those that stayed right ($R \to R$), those that stayed wrong ($W \to W$), and those that flip from right to wrong ($R \to W$). For Llama, the average margin for newly correct responses increases from $0.040$ to $0.072$, indicating robustness gains for knowledge that improved correctness. On the other hand, the margin for examples that either remained incorrect or flipped from correct to incorrect decreased slightly. We note that due to the sample size of $R\to W$ being so small, the result is not significant enough to make a generalization. Qwen and Gemma show similar results, where the margin for newly correct responses increases significantly. For all models, the margin for examples that remained correct increases after training. However, training slightly increases confidence for wrong answers in Qwen and Gemma. Our results show that knowledge injection can improve mechanistic uncertainty through increased average causal robustness margins.

\section{Conclusion}

We introduced mechanistic uncertainty as the fragility of the internal computation supporting an LLM’s response. Using the JLens, we implement it through the causal robustness margin $\alpha^*$, the minimum intervention to semantically deviate from the baseline response. Experiments show incorrect responses generally exhibit smaller margins, and $\alpha^*$ complements output-based uncertainty measures in predicting correctness. A controlled knowledge injection experiment shows increased margins for newly correct responses. These findings support causal robustness as a mechanistic uncertainty signal that connects to predictive reliability.

\section*{Acknowledgments}
This work is supported by U.S. National Library of Medicine, National Institutes of Health, under award number R00LM014308.

\section*{Limitations}
Since our focus in this work is mechanistic in nature, we evaluate a limited set of open-source LLMs where we assume complete white-box access, including hidden states and token probabilities. Furthermore, the proposed causal robustness margin is based on Jlens-derived insights, including Jlens vector steer direction, layer bands, and normalization, which could vary for different models. Furthermore, the method requires sweeping several $\alpha$ values and apply steers across a whole band, which can be computationally expensive. However, our main focus in this work is interpretability rather than efficiency. Despite these constraints, we show that the causal robustness margin is an uncertainty signal that consistently compliments conventional output-centric uncertainty methods and can reveal internal uncertainty behaviors in LLMs.

\section*{Ethical Considerations}
This paper intends to advance the areas of interpretability and uncertainty quantification for language models and to make LLMs more reliable and trustworthy. All models used in this work are open-source. The datasets used are all publicly available.

\bibliography{main}

\clearpage
\appendix

\clearpage

\section*{Appendix}
\label{sec:appendix}
\section{Dataset details}

\paragraph{TriviaQA \cite{joshi-etal-2017-triviaqa}} TriviaQA is an open-domain factual question answering dataset collected from trivia and quiz league websites. Each question is associated with one or more semantically equivalent ground-truth answers. We randomly sample $1000$ examples from the validation set for our experiments.

\paragraph{SciQ \cite{welbl-etal-2017-sciq}} SciQ is a science question answering dataset consisting of crowdsourced multiple-choice questions covering topics such as physics, chemistry, and biology. Each question contains four answer choices with a single correct answer. The dataset contains $13{,}679$ questions from which we randomly sample $1000$ for our experiment. We do not include the answer choices in the query for our experiments.

\paragraph{TruthfulQA \cite{lin-etal-2022-truthfulqa}} TruthfulQA is a question answering benchmark designed to evaluate whether language models produce truthful responses rather than reproduce common misconceptions or false beliefs. It contains $817$ questions spanning $38$ categories, including health, law, finance, and politics. Each question is accompanied by reference truthful and false answers, enabling evaluation of generated responses for both truthfulness and informativeness. We utilize the whole dataset for our experiment.

\paragraph{MedQA \cite{medqa}.} MedQA is a medical question answering benchmark derived from professional medical licensing examinations. We use the English USMLE subset, which consists of $12{,}723$ questions covering medical knowledge and clinical reasoning from which we randomly sample $1000$ for our experiment. Each question is presented in a multiple-choice format with a single correct answer. 

\paragraph{PopQA \cite{mallen-etal-2023-popqa}} PopQA is an entity-centric open-domain factual question answering dataset containing approximately $14{,}000$ question--answer pairs derived from Wikidata. We randomly sample $1000$ examples for our experiment. The dataset associates entities with Wikipedia page-view statistics, providing a measure of entity popularity. PopQA emphasizes long-tail and less popular entities for which language models typically exhibit weaker factual recall, making it useful for studying the relationship between parametric knowledge and model confidence.

\begin{figure}[t]
    \centering
    \begin{tcolorbox}[
        width=\columnwidth,
        title={Contextualization Prompt},
        colback=gray!5,
        colframe=gray!60,
        fonttitle=\bfseries,
        fontupper=\small,
        boxrule=0.5pt,
        arc=2pt
    ]
    You are provided with a factual statement about the following
    subject: `\texttt{\{subject\}}'. Specifically, the factual
    statement focuses on the following entity:
    `\texttt{\{subject\}}' within the context of the subject.
    The statement was taken out of the passage below and does
    not name its subject, so on its own it is unclear who or
    what it is about. Rewrite it so that it explicitly names
    `\texttt{\{subject\}}', resolving pronouns and references
    such as `he', `she', `it', `the film' or `the musical',
    in a way that fits naturally within the sentence.
    Keep every fact exactly as stated and add no other
    information. Return just the contextualized statement
    and nothing else.

    \medskip
    \noindent Statement: \texttt{\{fact\}}

    \medskip
    \noindent The passage about \texttt{\{subject\}}, for reference:

    \noindent\texttt{\{passage\}}
    \end{tcolorbox}
    \caption{Prompt for contextualizing with Qwen3-32B}
    \label{fig:contextualization_prompt}
\end{figure}

\section{Task-agnostic Instruction SFT details}
\label{app:sft}
For our knowledge injection experiment on the PopQA dataset, we follow the steps outlined in the Knowledge-Instruct paper \cite{ovadia2025knowledge}. The curated PopQA dataset consists of 1,935 long-tail knowledge question and answer pairs, each with a corresponding chunk of context retrieved from Wikipedia. We start with this context and build a training dataset suitable for instruction tuning for knowledge injection with $\mathcal{M}_{ext}$ as an extraction/summarization LLM (Qwen3-32B).

\begin{itemize}
    \item \textbf{Contextualization}. First, we split the context into separate facts by treating each sentence as a fact and ensure that the subject appears in each fact using $\mathcal{M}_{ext}$.
    \item \textbf{Deduplication}. Within each factual set, we remove redundant facts using $\mathcal{M}_{ext}$.
    \item \textbf{Paraphrasing}. For each fact, we generate a set of $5$ paraphrased variations using $\mathcal{M}_{ext}$.
    
    \item \textbf{Instruction Conversion}. We then convert each paraphrased fact into an instruction response pair (instruction, fact), where `instruction' is a generic instruction template contextualized with the subject sampled from a list of 25 templates specified in \cite{ovadia2025knowledge} including phrases such as \textit{``Tell me a fact about \{entity\}''}, \textit{``What can you tell me about \{entity\}''}, etc..
\end{itemize}
We perform supervised finetuning (SFT) on a total of 126,364 pairs where 10,000 of them are generic instruct-response pairs from the OpenOrca \cite{OpenOrca} dataset to ensure the model remains faithful to its instruction following. We perform LoRA training with rank $64$ and learning rate 1e-4 for 2 epochs on all models.

\begin{figure}[t]
    \centering
    \begin{tcolorbox}[
        width=\columnwidth,
        title={Paraphrasing Prompt},
        colback=gray!5,
        colframe=gray!60,
        fonttitle=\bfseries,
        fontupper=\small,
        boxrule=0.5pt,
        arc=2pt
    ]
    Paraphrase the following factual statement \texttt{\{k\}} times.
    Each paraphrase must keep exactly the same meaning and every
    specific detail: do not change proper names, titles, dates,
    numbers or locations, and do not add or remove any information.
    Each paraphrase must be a complete standalone sentence that
    names `\texttt{\{subject\}}', and the paraphrases should differ
    from the original and from each other in wording and sentence
    structure. Return only a JSON array of \texttt{\{k\}} strings
    and nothing else.

    \medskip
    \noindent Statement: \texttt{\{fact\}}
    \end{tcolorbox}
    \caption{Prompt for generating paraphrases with Qwen3-32B}
    \label{fig:paraphrasing_prompt}
\end{figure}

\begin{table}[t]
\centering
\small
\setlength{\tabcolsep}{4pt}
\caption{
Baseline accuracy and recovery after steering at the causal
robustness margin $\alpha^*$.
Recovered is the percentage of initially incorrect responses
made correct. Oracle-steered accuracy preserves initially correct
responses and steers only incorrect responses, requiring
ground-truth correctness.
}
\label{tab:steering_recovery}

\resizebox{\columnwidth}{!}{%
\begin{tabular}{llccc}
\toprule
\textbf{Dataset}
& \textbf{Model}
& \shortstack{\textbf{Baseline}\\\textbf{Acc.}}
& \shortstack{\textbf{Recovered}\\\textbf{(\%)}}
& \shortstack{\textbf{Oracle}\\\textbf{Acc.}} \\
\midrule
\multirow{3}{*}{TruthfulQA}
& Llama-3.1-8B & 50.6\% & 24.8\% & 62.8\% \\
& Qwen2.5-7B   & 64.5\% & 19.7\% & 71.5\% \\
& Gemma-3-12B  & 57.5\% & 17.3\% & 64.9\% \\
\midrule
\multirow{3}{*}{TriviaQA}
& Llama-3.1-8B & 72.8\% & 24.6\% & 79.5\% \\
& Qwen2.5-7B   & 55.9\% & 16.8\% & 63.3\% \\
& Gemma-3-12B  & 72.3\% & 12.3\% & 75.7\% \\
\midrule
\multirow{3}{*}{MedQA}
& Llama-3.1-8B & 67.6\% & 52.2\% & 84.5\% \\
& Qwen2.5-7B   & 55.4\% & 37.9\% & 72.3\% \\
& Gemma-3-12B  & 62.7\% & 19.0\% & 69.8\% \\
\midrule
\multirow{3}{*}{PopQA}
& Llama-3.1-8B & 31.5\% & 6.3\% & 35.8\% \\
& Qwen2.5-7B   & 19.1\% & 6.1\% & 24.0\% \\
& Gemma-3-12B  & 25.3\% & 2.8\% & 27.4\% \\
\midrule
\multirow{3}{*}{SciQ}
& Llama-3.1-8B & 53.7\% & 20.3\% & 63.1\% \\
& Qwen2.5-7B   & 52.0\% & 16.0\% & 59.7\% \\
& Gemma-3-12B  & 59.2\% & 11.3\% & 63.8\% \\
\bottomrule
\end{tabular}%
}
\end{table}

\section{Additional Experiments}
\label{sec:app_examples}

\paragraph{Oracle Correctness after Steering.} We also report an oracle steering experiment, where we apply the causal robustness margin $\alpha^*$ steer for only incorrect answers to observe how many answers are recoverable after a Jlens steer. The results are shown in \cref{tab:steering_recovery}. We observe that for most datasets, we can recover up to $10\%$ - $20\%$ of answers from the steer. We find that for many of these questions, the Jspace consensus reads the correct answers, but the model verbalizes an incorrect answer. This behavior illustrates that even for questions the model is uncertain about, factually correct answers may reside close in the Jspace.

\begin{figure*}[t]
    \centering

    \begin{subfigure}{\textwidth}
        \centering
        \includegraphics[width=\linewidth]{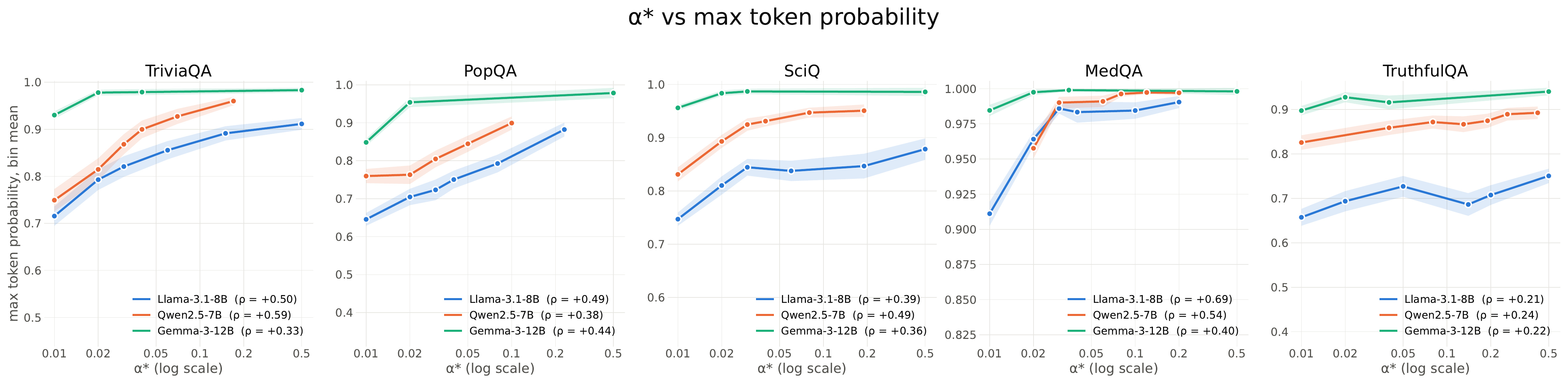}
        \caption{Max token probability}
        \label{fig:alpha-seq-likelihood}
    \end{subfigure}

    \medskip

    \begin{subfigure}{\textwidth}
        \centering
        \includegraphics[width=\linewidth]{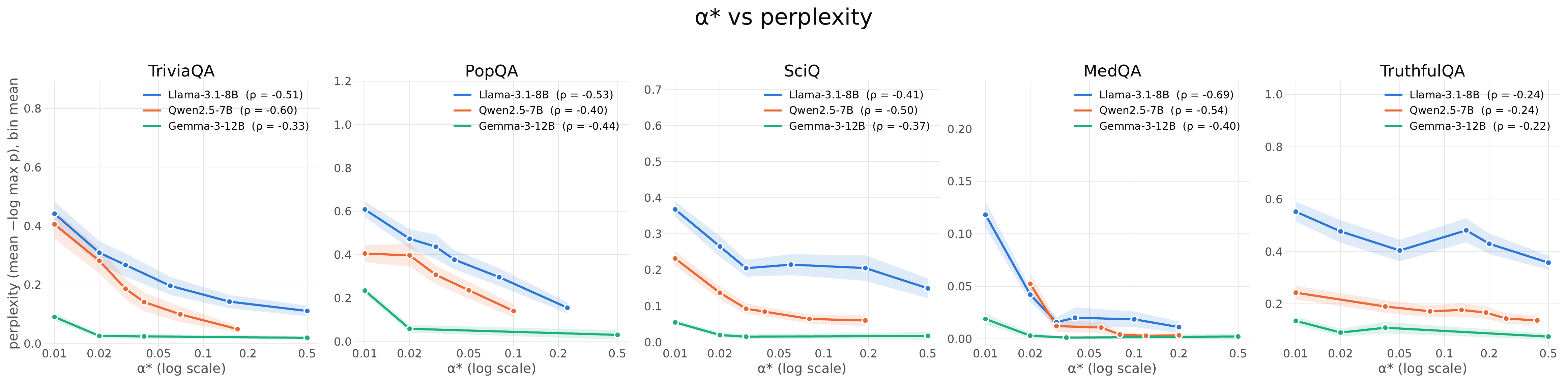}
        \caption{Perplexity}
        \label{fig:alpha-perplexity}
    \end{subfigure}

    \medskip

    \begin{subfigure}{\textwidth}
        \centering
        \includegraphics[width=\linewidth]{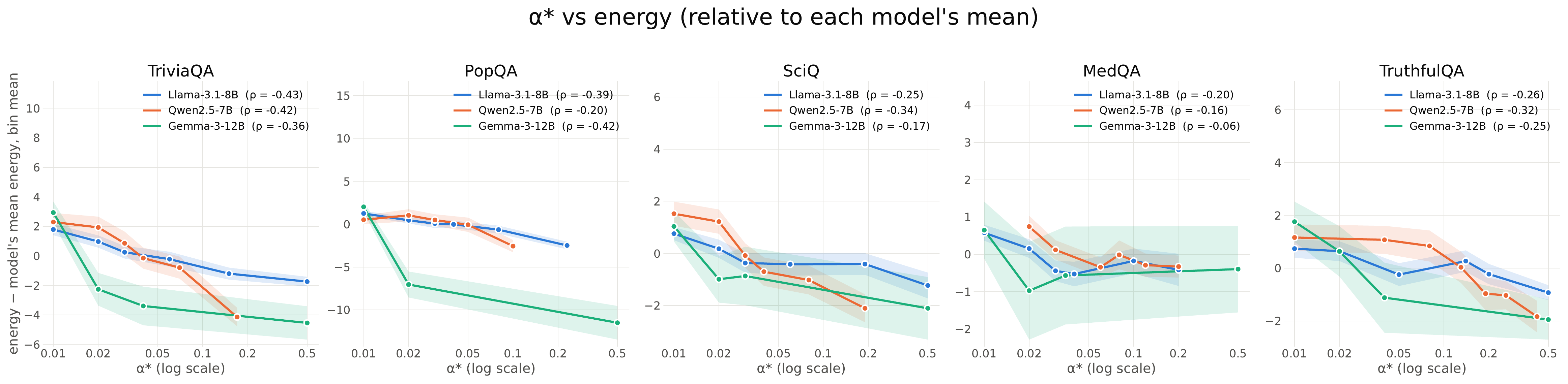}
        \caption{Energy}
        \label{fig:alpha-energy}
    \end{subfigure}

    \medskip

    \begin{subfigure}{\textwidth}
        \centering
        \includegraphics[width=\linewidth]{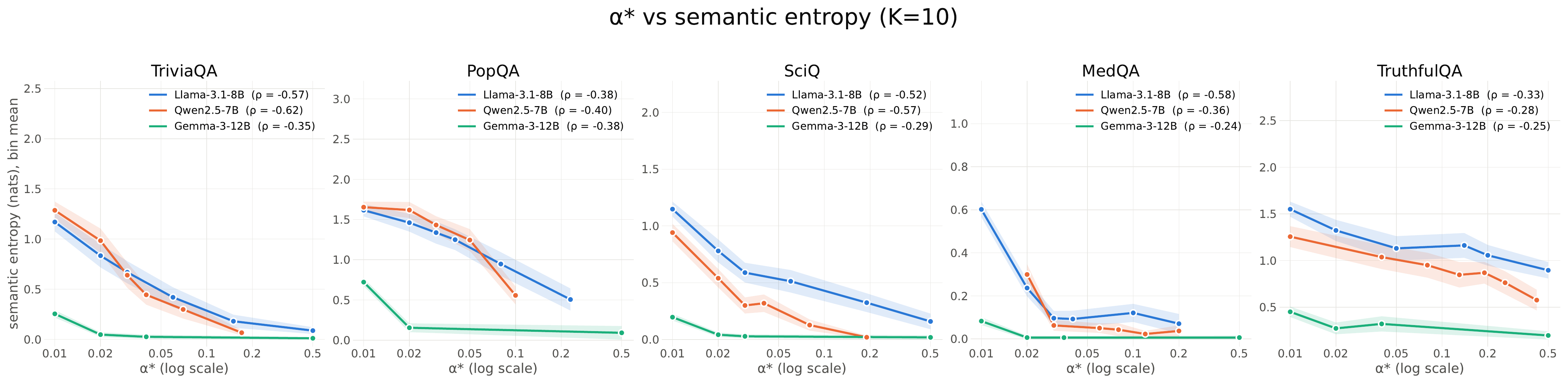}
        \caption{Semantic entropy}
        \label{fig:alpha-semantic-entropy}
    \end{subfigure}

    \caption{
    Correlation between the causal robustness margin $\alpha^*$
    and sequence likelihood, perplexity, energy, and semantic entropy
    across models and datasets.
    }
    \label{fig:alpha-uncertainty-signals}
\end{figure*}

\paragraph{Mechanistic Uncertainty Correlation with other Uncertainty Signals.} In \cref{fig:alpha-uncertainty-signals}, we include correlations with several additional uncertainty metrics such as max token probability, perplexity, energy \cite{liu2020energy}, and semantic entropy. Max token probability is computed by taking the average of the maximum probability of the token distribution at each generated token. Semantic entropy \cite{kuhn2023semantic} is computed over $K=10$ completions with temperature $T=0.7$, grouped into semantic clusters using an NLI model or string match, and computing entropy over the clusters. We find that $\alpha^*$ is positively correlated with max token probability and negatively correlated with perplexity, energy, and semantic entropy. As with self-consistency and mean token entropy, the strongest correlations are often exhibited in the TriviaQA dataset.

\begin{figure*}[t]
    \centering
    \begin{subfigure}[t]{\textwidth}
        \centering
        \includegraphics[width=\linewidth]{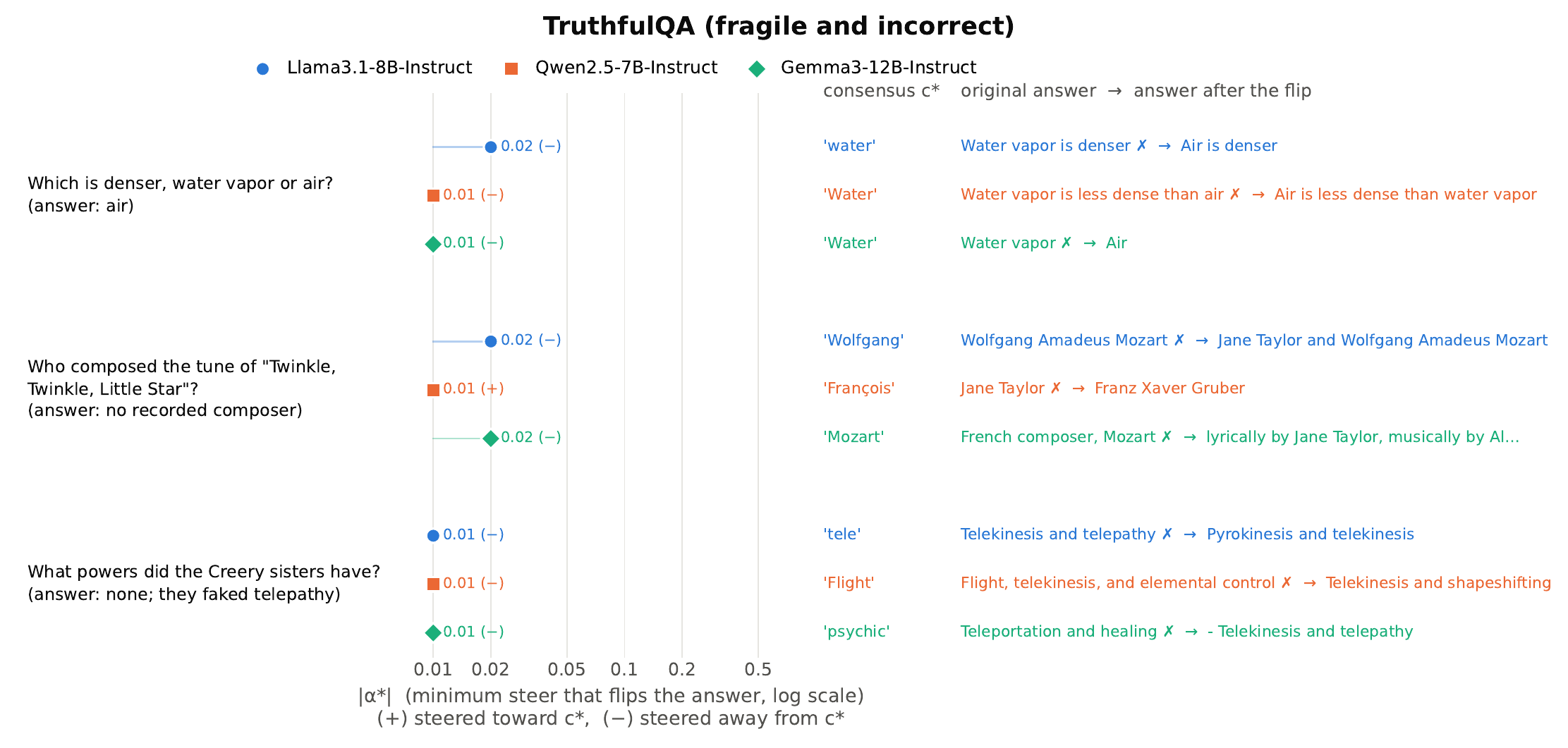}
        \caption{}
        \label{fig:truthful_ex_a}
    \end{subfigure}
    \hfill
    \vspace{3cm}
    \begin{subfigure}[t]{\textwidth}
        \centering
        \includegraphics[width=\linewidth]{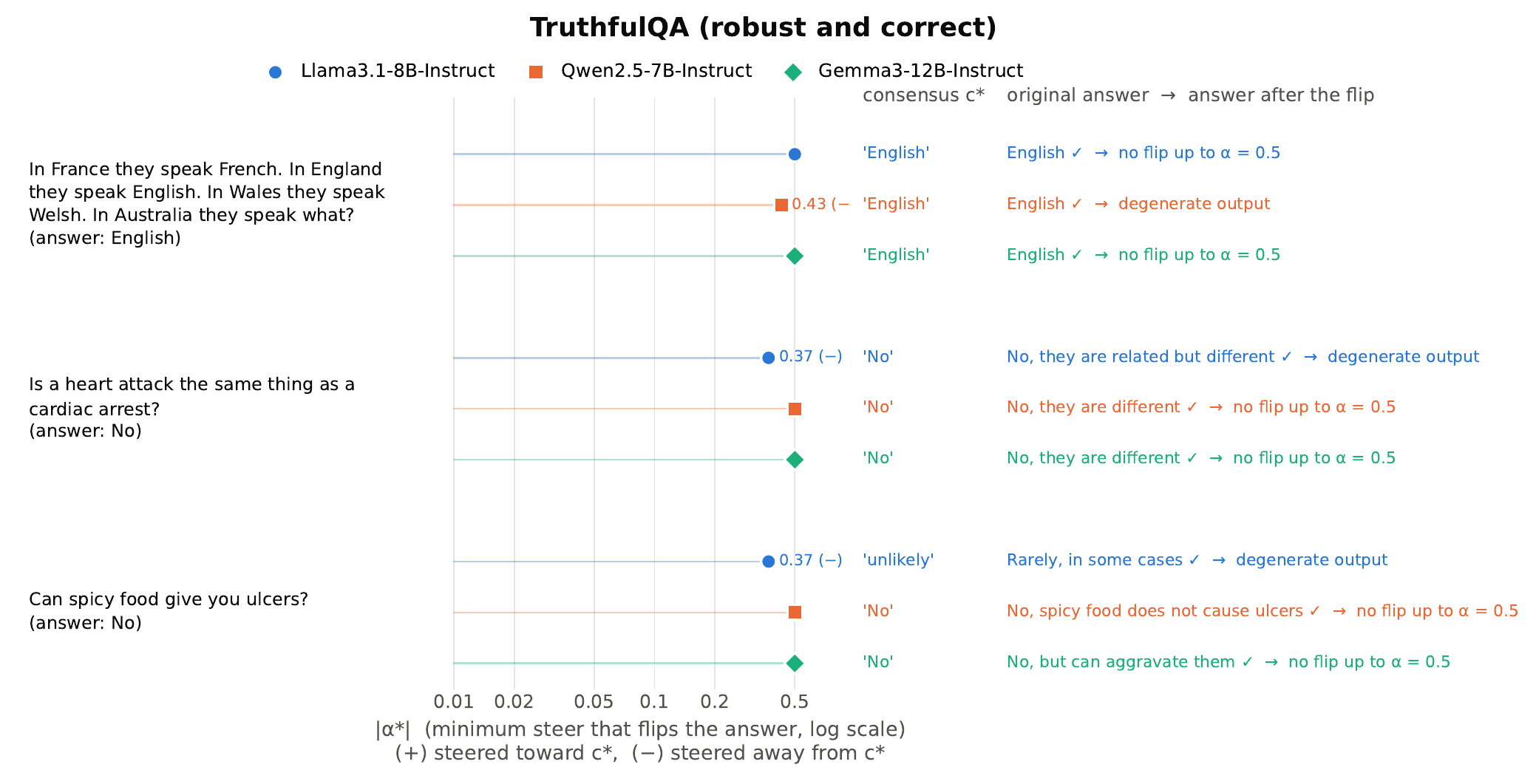}
        \caption{}
        \label{fig:truthful_ex_b}
    \end{subfigure}
    \caption{TruthfulQA flip examples.}
    \label{fig:truthful_ex}
\end{figure*}

\begin{figure*}[t]
    \centering
    \begin{subfigure}[t]{\textwidth}
        \centering
        \includegraphics[width=\linewidth]{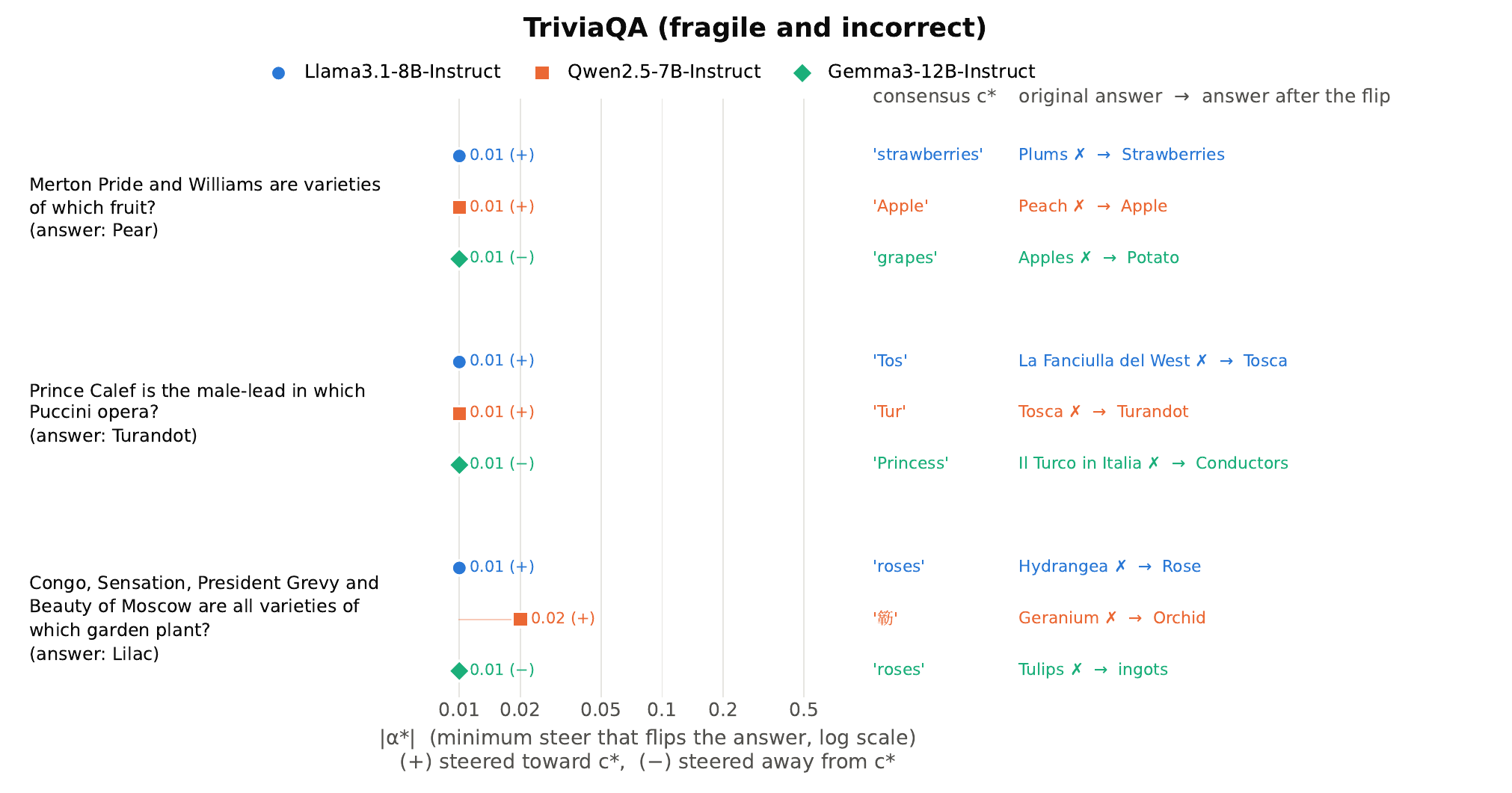}
        \caption{}
        \label{fig:trivia_ex_a}
    \end{subfigure}
    \hfill
    \vspace{3cm}
    \begin{subfigure}[t]{\textwidth}
        \centering
        \includegraphics[width=\linewidth]{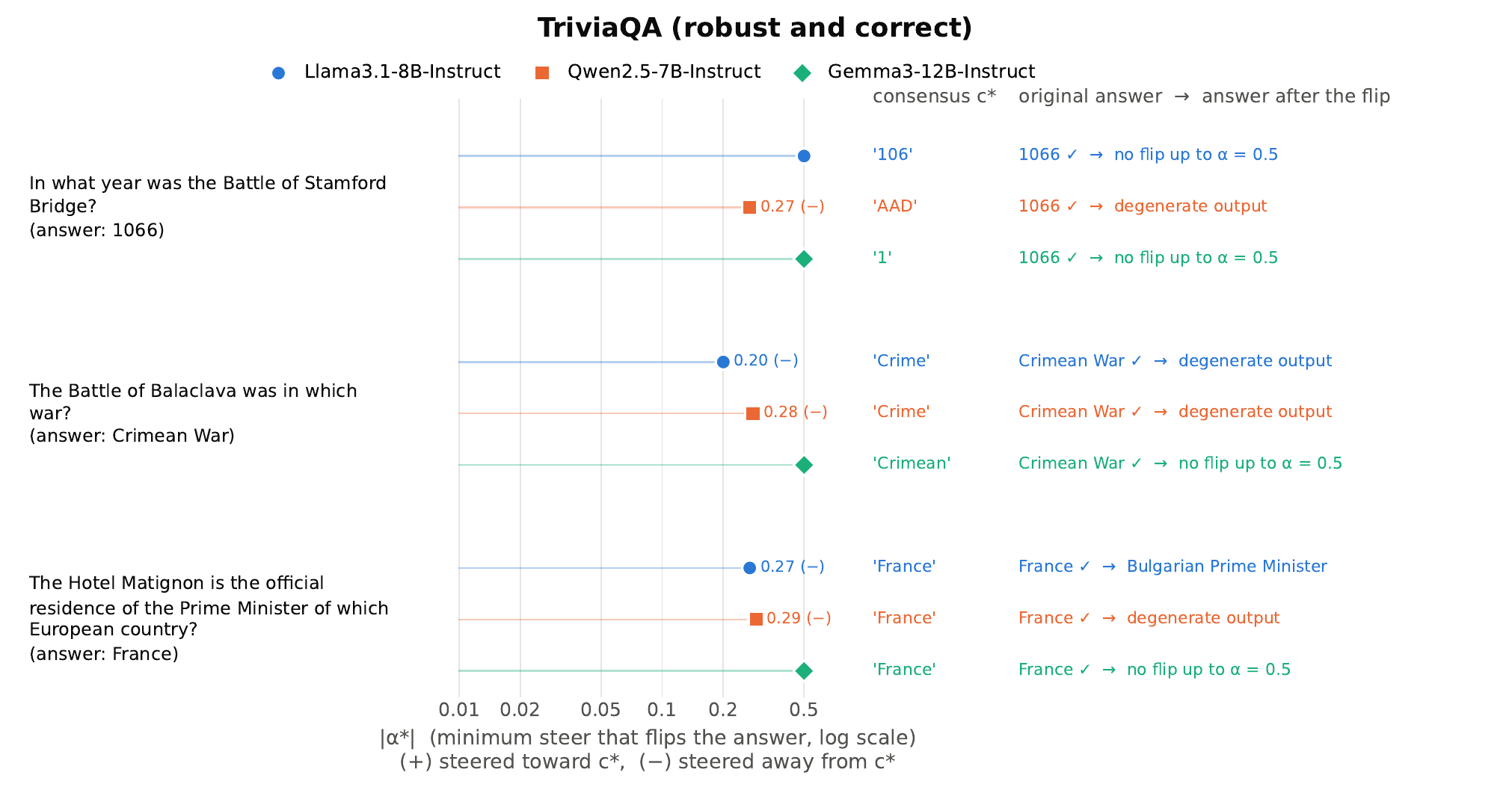}
        \caption{}
        \label{fig:trivia_ex_b}
    \end{subfigure}
    \caption{TriviaQA flip examples.}
    \label{fig:trivia_ex}
\end{figure*}

\begin{figure*}[t]
    \centering
    \begin{subfigure}[t]{\textwidth}
        \centering
        \includegraphics[width=\linewidth]{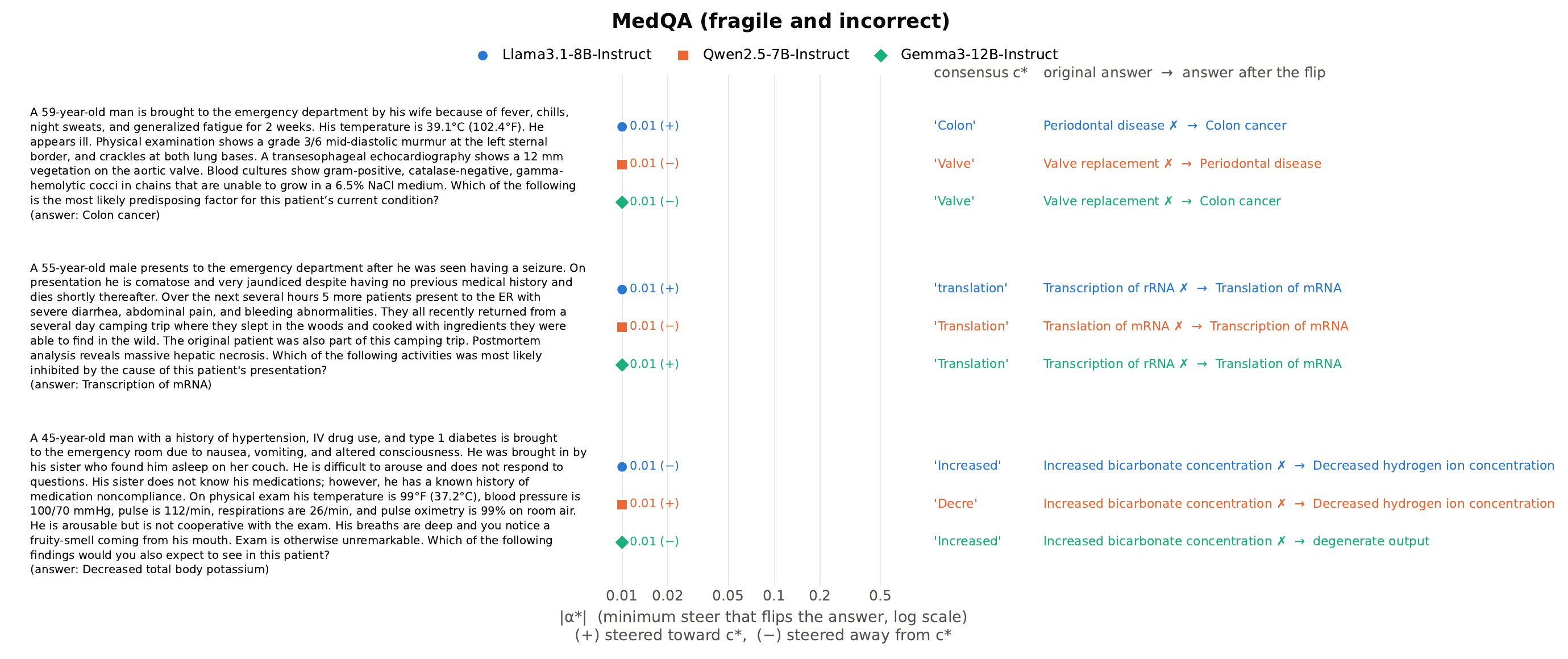}
        \caption{}
        \label{fig:medqa_ex_a}
    \end{subfigure}
    \hfill
    \vspace{3cm}
    \begin{subfigure}[t]{\textwidth}
        \centering
        \includegraphics[width=\linewidth]{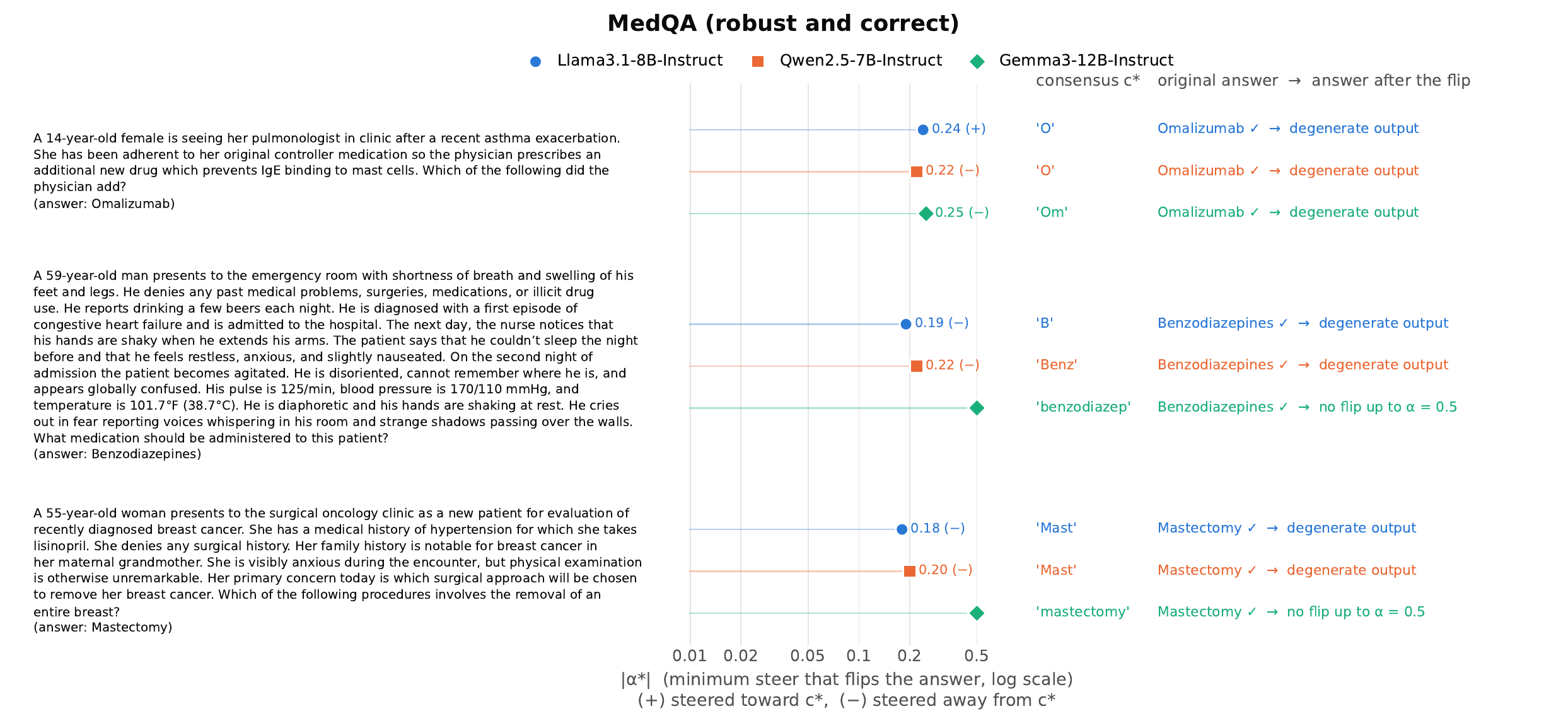}
        \caption{}
        \label{fig:medqa_ex_b}
    \end{subfigure}
    \caption{MedQA flip examples.}
    \label{fig:medqa_ex}
\end{figure*}

\begin{figure*}[t]
    \centering
    \begin{subfigure}[t]{\textwidth}
        \centering
        \includegraphics[width=\linewidth]{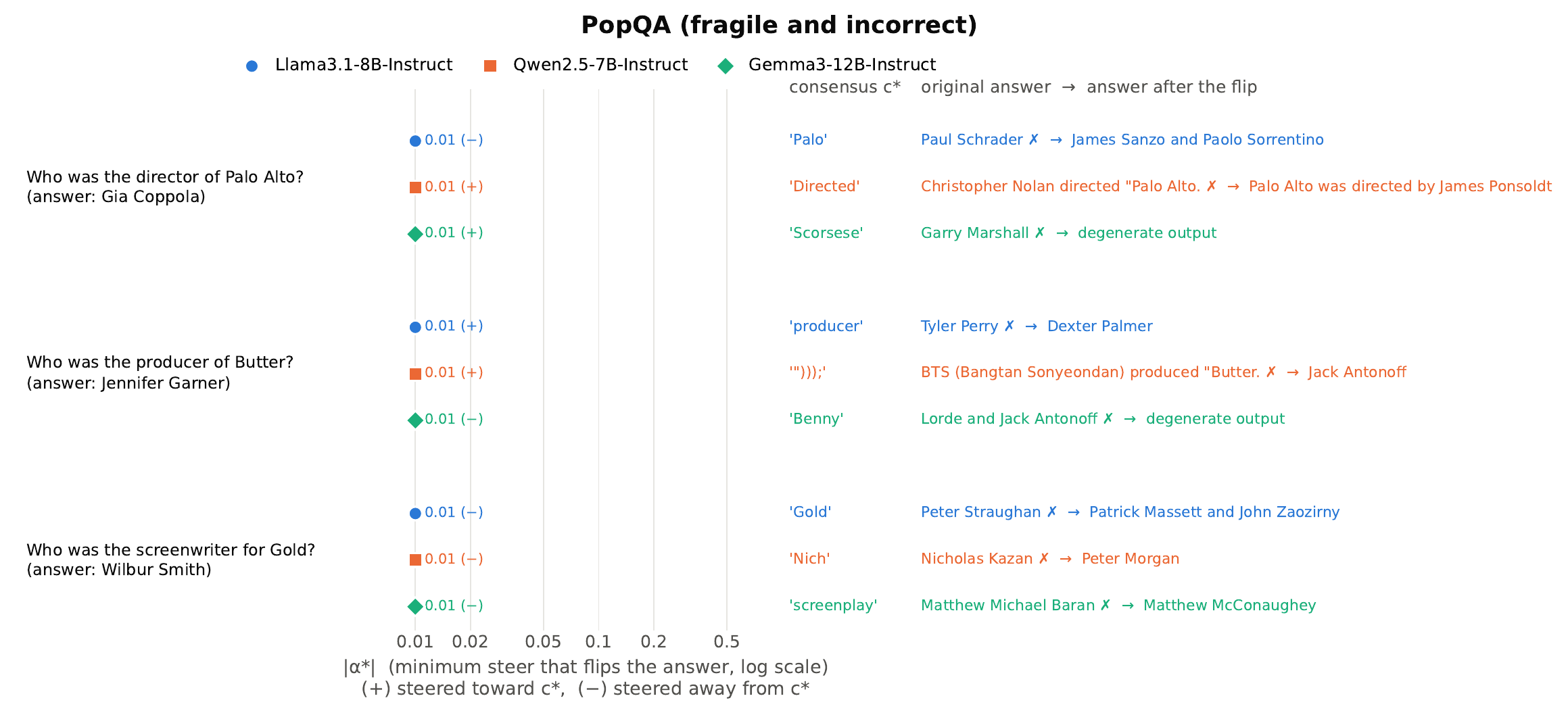}
        \caption{}
        \label{fig:popqa_ex_a}
    \end{subfigure}
    \hfill
    \vspace{3cm}
    \begin{subfigure}[t]{\textwidth}
        \centering
        \includegraphics[width=\linewidth]{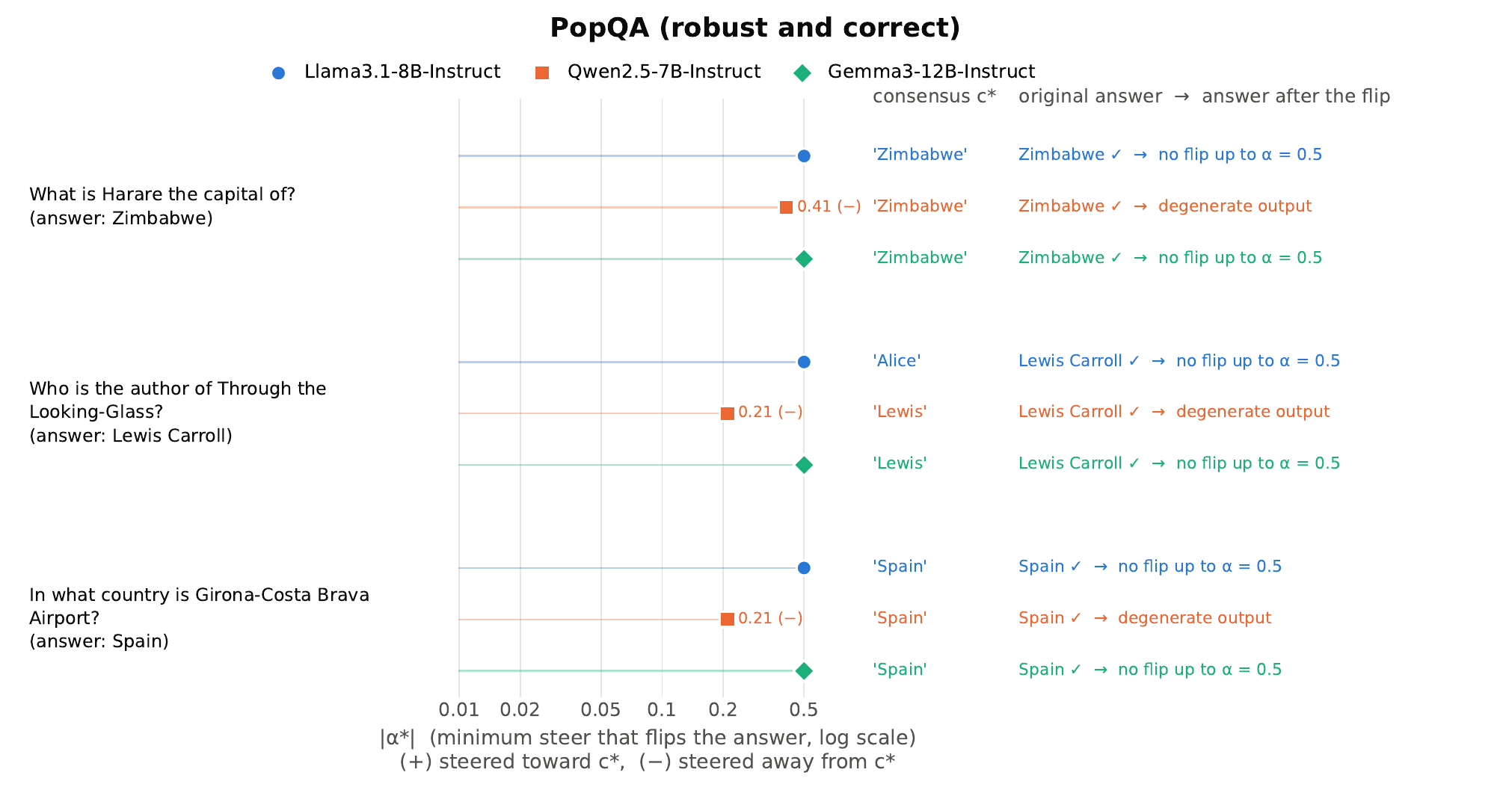}
        \caption{}
        \label{fig:popqa_ex_b}
    \end{subfigure}
    \caption{PopQA flip examples.}
    \label{fig:popqa_ex}
\end{figure*}

\begin{figure*}[t]
    \centering
    \begin{subfigure}[t]{\textwidth}
        \centering
        \includegraphics[width=\linewidth]{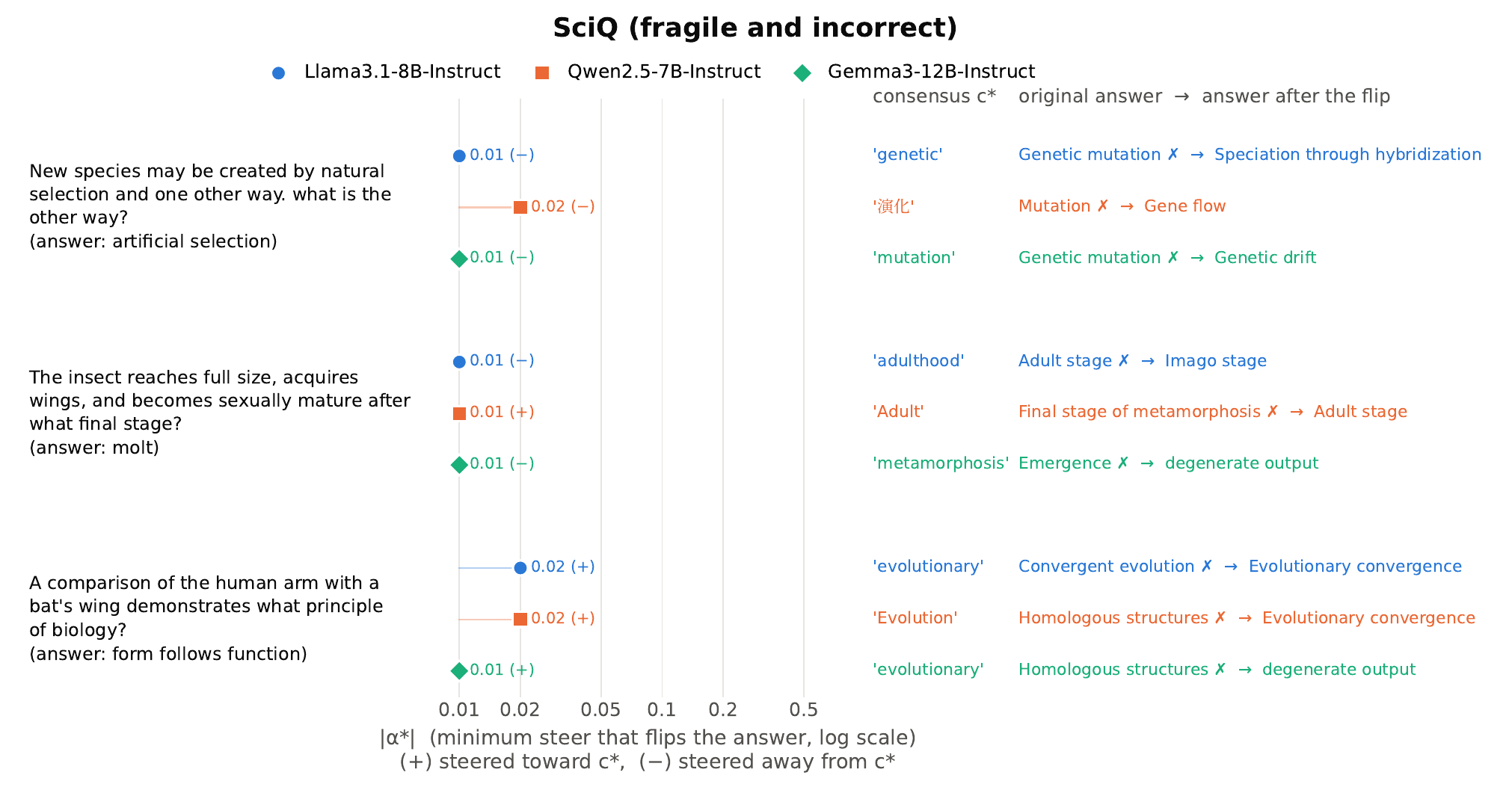}
        \caption{}
        \label{fig:sciq_ex_a}
    \end{subfigure}
    \hfill
    \vspace{3cm}
    \begin{subfigure}[t]{\textwidth}
        \centering
        \includegraphics[width=\linewidth]{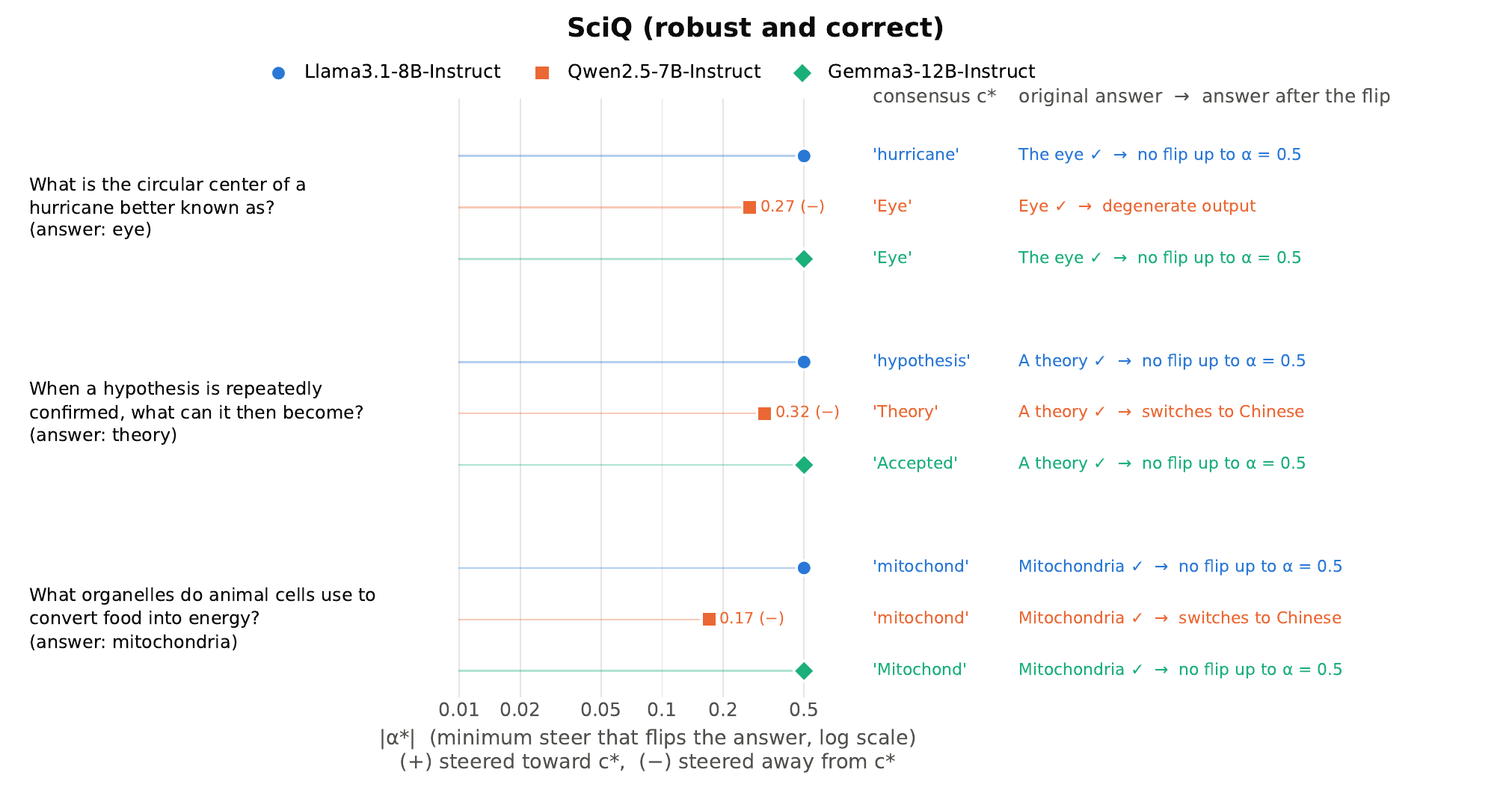}
        \caption{}
        \label{fig:sciq_ex_b}
    \end{subfigure}
    \caption{SciQ flip examples.}
    \label{fig:sciq_ex}
\end{figure*}

\paragraph{Intuitive Examples of Mechanistic Uncertainty.} In \cref{fig:truthful_ex,fig:trivia_ex,fig:medqa_ex,fig:popqa_ex,fig:sciq_ex}, we include several examples from each dataset of fragile/incorrect and robust/correct responses to illustrate how the causal robustness margin captures uncertainty through mechanism stability.

\section{Computational Resources}
 We mainly utilize a workstation with an NVIDIA RTX PRO 6000 with 96GB RAM, and 2 NVIDIA H100 GPUs with 94GB RAM for our experiments.

\end{document}